\documentclass[letterpaper, 10 pt, conference]{ieeeconf}  % Comment this line out if you need a4paper

\IEEEoverridecommandlockouts                              % This command is only needed if 
\usepackage[T1]{fontenc}
\usepackage{graphicx}
\usepackage{epstopdf}
\usepackage{bbm}
\usepackage{amsfonts}
\usepackage{amsmath}
\usepackage{multirow}
\usepackage{booktabs}
\usepackage{bm}
\usepackage{hyperref}
\usepackage{bbding}
\usepackage{float}
\usepackage{stfloats}
\usepackage{cite}

\title{\LARGE \bf
Leveraging Semantic and Geometric Information for Zero-Shot Robot-to-Human Handover
}

\author{Jiangshan Liu, Wenlong Dong, Jiankun Wang$^{*}$, \textit{Senior Member, IEEE}, Max Q.-H. Meng$^{*}$, \textit{Fellow, IEEE}
\thanks{This work was partly supported by Shenzhen Key Laboratory of Robotics
Perception and Intelligence (ZDSYS20200810171800001) and the Shenzhen Science and Technology Program under Grant RCBS20221008093305007, 20231115141459001, Young Elite Scientists Sponsorship Program by CAST under Grant 2023QNRC001, High level of special funds (G03034K003) from Southern University of Science and Technology, Shenzhen, China. (Corresponding author: Jiankun Wang).}
\thanks{J. Liu, W. Dong, J. Wang and M. Meng are with Shenzhen Key
Laboratory of Robotics Perception and Intelligence and the Department of
Electronic and Electrical Engineering, Southern University of Science and
Technology, Shenzhen, China. J. Wang is also with the Jiaxing Research
Institute, Southern University of Science and Technology, Jiaxing, China.}% <-this % stops a space
\thanks{$^{*}$Corresponding authors: Jiankun Wang, Max Q.-H. Meng. e-mail:
wangjk@sustech.edu.cn, max.meng@ieee.org.
}
}

\begin{document}

\maketitle
\thispagestyle{empty}
\pagestyle{empty}

%%%%%%%%%%%%%%%%%%%%%%%%%%%%%%%%%%%%%%%%%%%%%%%%%%%%%%%%%%%%%%%%%%%%%%%%%%%%%%%%
\begin{abstract}
% Human-robot interaction (HRI) encompasses a wide range of collaborative tasks, with handover being one of the most fundamental. As robots become more integrated into human environments, the potential for service robots to assist in handing objects to humans is becoming increasingly promising. In robot-to-human (R2H) handover, selecting the optimal grasp is vital for a successful handover, which requires avoiding interference with the human’s preferred grasp region and minimizing intrusion into their workspace. Existing methods either inadequately consider geometric information or rely on data-driven approaches, which often struggle to generalize across diverse objects. To address these limitations, we propose a novel zero-shot system that combines semantic and geometric information to generate optimal handover grasps. We validate our approach through ablation studies, and real-world comparison experiments. Results demonstrate that our system improves handover success rates and provides a more user-preferred interaction experience.
Human-robot interaction (HRI) encompasses a wide range of collaborative tasks, with handover being one of the most fundamental. As robots become more integrated into human environments, the potential for service robots to assist in handing objects to humans is increasingly promising. In robot-to-human (R2H) handover, selecting the optimal grasp is crucial for success, as it requires avoiding interference with the human's preferred grasp region and minimizing intrusion into their workspace. Existing methods either inadequately consider geometric information or rely on data-driven approaches, which often struggle to generalize across diverse objects. To address these limitations, we propose a novel zero-shot system that combines semantic and geometric information to generate optimal handover grasps. Our method first identifies grasp regions using semantic knowledge from vision-language models (VLMs) and, by incorporating customized visual prompts, achieves finer granularity in region grounding. A grasp is then selected based on grasp distance and approach angle to maximize human ease and avoid interference. We validate our approach through ablation studies and real-world comparison experiments. Results demonstrate that our system improves handover success rates and provides a more user-preferred interaction experience. Videos, appendixes and more are available at \url{https://sites.google.com/view/vlm-handover/}.
% This electronic document is a ÒliveÓ template. The various components of your paper [title, text, heads, etc.] are already defined on the style sheet, as illustrated by the portions given in this document.

\end{abstract}

%%%%%%%%%%%%%%%%%%%%%%%%%%%%%%%%%%%%%%%%%%%%%%%%%%%%%%%%%%%%%%%%%%%%%%%%%%%%%%%%
\section{INTRODUCTION}
\label{intro}
Human-robot interaction is a consistently popular research area, drawing attention for its potential to improve human-robot collaboration. Human-robot handover is a vital and practical task in human-robot collaboration as more and more robots are integrated into human workspaces \cite{duan2024human}. In particular, robot-to-human (R2H) handover has real-world applications across diverse scenarios, from handing tools to workers in factory \cite{castro2021trends,wang2021survey, xu2023learning} to fetching daily objects for people at home \cite{rahman2016trust}, or even robot nurse to assist surgery at hospitals \cite{ortenzi2021object,wang2020neural, ortenzi_object_2022}. A handover typically consists of two phases: the pre-handover grasp and the delivery, during which interaction between the human and robot occurs, making the selection of the pre-handover grasp critical. 
% In this work, we leverage semantic and geometric information in the grasp generation algorithm to enhance handover performance. 

\begin{figure}[htbp]
\centering
\setlength{\abovecaptionskip}{0.cm}
\includegraphics[width=1.0\linewidth]{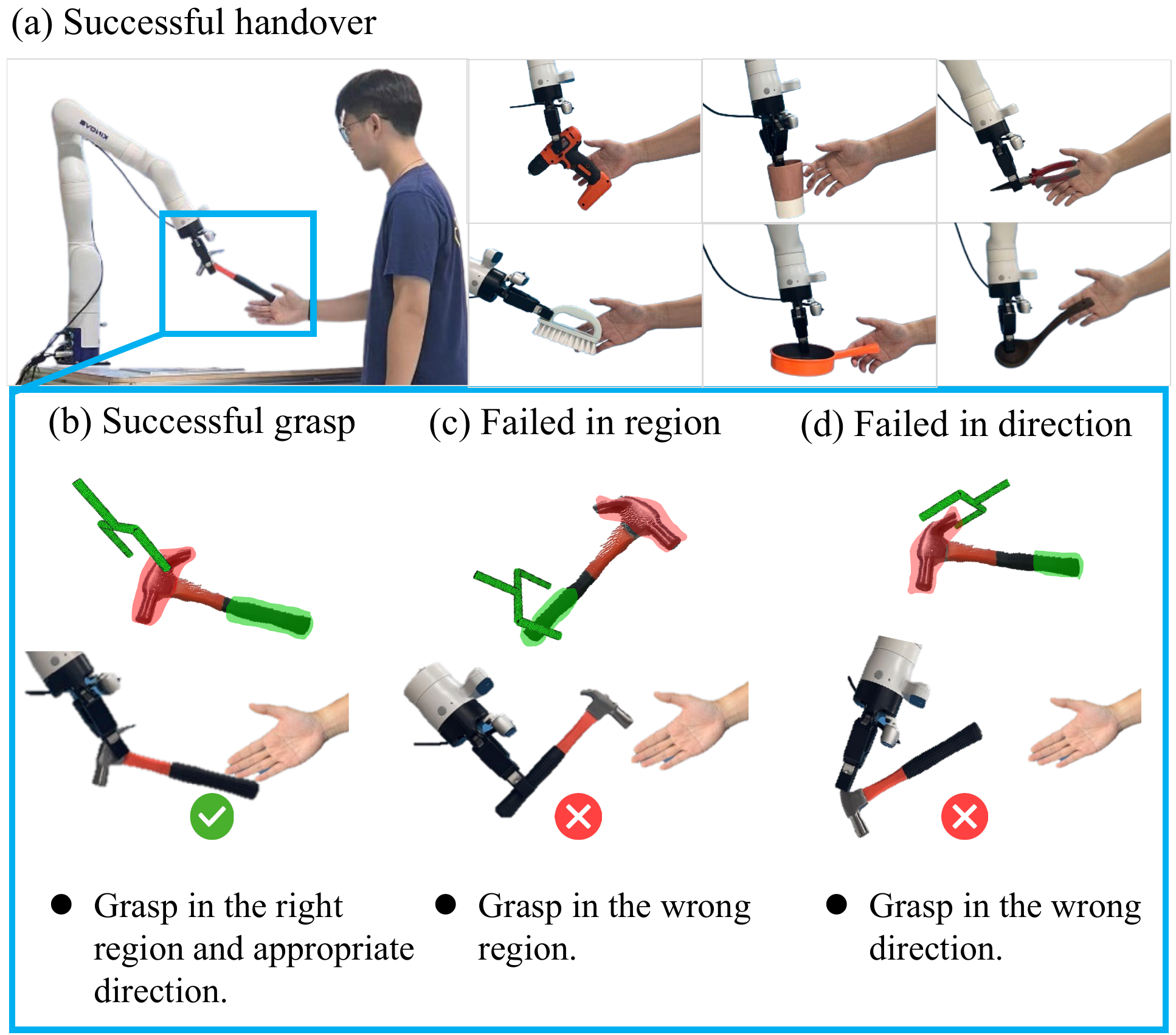}
\caption{(a) Successful robot-to-human handovers share the same pattern in which humans tend to grasp regions that are conducive to the object's intended function. (b) A successful grasp for handover takes into account the robot's grasp region and direction. Red and green masks are predicted regions where robots and humans grasp. (c) A grasp fails in handover for grasping on region that humans prefer to grasp. (d) A grasp fails in handover for grasping from an inappropriate direction by intruding humans' workspace. }
\label{head}
\vspace{-0.1in}
\end{figure}

Two key factors are essential for a successful handover. First, the robot must grasp the object while ensuring sufficient space is left for the human to comfortably and efficiently use it. Care must also be taken to avoid the human's preferred grasping area, typically the part held during the object's intended use as Fig. \ref{head}(b). Second, the robot should consider the grasping direction to minimize intrusion into the human's workspace. For example, the second subplot in Fig. \ref{head}(c) illustrates a failed handover caused by an inappropriate direction of grasp.  
% Furthermore, it's important to deliver the object to a position and orientation that make it both accessible and easy to use for subsequent tasks. Secondly, it matters how to deliver the object to a position and orientation where it is both accessible and easy for use at subsequent tasks.

Existing methods have attempted to solve these by leveraging the semantic information of objects. Some approaches perform semantic segmentation \cite{christensen2022learning, ortenzi_robot_2022} or predict human contact points \cite{wang_contacthandover_2024} to guide grasp selection. However, they either inadequately consider geometric information during the handover process or rely on data-driven approaches that lead to poor generalization. Recently, some methods have leveraged the strong generalization and reasoning capabilities of foundation models \cite{liang2023code,huang2023voxposer}, including large language models (LLMs) and vision-language models (VLMs), in robotic manipulation tasks. Works like LAN-grasp \cite{mirjalili2023lan} directly utilize foundation models' visual grounding ability to perform handover tasks by grounding grasp regions. However, it only provides coarse information about the object parts and does not constrain the grasp direction. 

To improve handover performance further and be applicable to a more diverse range of objects, we propose a zero-shot R2H object handover system that utilizes semantic prior in VLMs to ground the most possible interaction regions on objects. Meanwhile, we constrain the grasp direction using geometric information to better adapt to the handover process.

% Inspired by LAN-grasp, we introduce VLM to robot-human handover task to help extract object part-level semantics.

As illustrated in Fig. \ref{figure}, our system consists of three modules. The first module, the handover region grounding module identifies where humans prefer to grasp during handover and determines the corresponding robot grasp regions. Next, the grasp selection module generates a range of grasp candidates and selects the most suitable grasp for the handover. Finally, the execution module calculates the optimal handover pose to minimize human effort and executes the grasp and subsequent delivery based on the human pose.

To validate the effectiveness of our proposed system, we performed a comparative analysis of the success rates for handover-associated grasps generated by our method versus baseline approaches. We also conducted ablation studies to assess the contribution of each component of our selecting algorithm. Finally, we verified the system's performance through real-world experiments. The main contributions of our paper are summarized as follows: 
\begin{itemize}
    % \item A vision-based robot-to-human handover system: Our system enables a robot to deliver objects to a human receiver by accounting for both human pose and object semantics.
    % \item A zero-shot handover grasp selection algorithm: This algorithm integrates semantic and geometric information to identify the optimal grasp for handover.
    % \item A validation of our method through various experiments and a user study: Our approach has been assessed through grasp selection and practical experiments with real robots, demonstrating that it achieves a higher success rate. User study showcases that our approach can generate handover grasps that are more preferred by users. 
    
    \item A zero-shot R2H handover method: This method integrates semantic and geometric information to identify the optimal grasp for handover.
    \item Leveraging the Set-of-Mark visual prompt to VLMs, we achieve finer granularity in grounding regions. To further refine grasp selection, we rank the potential grasps within regions by considering both distance and approach angle.
    \item A vision-based robot-to-human handover system: Our system is able to choose the best grasp to handover objects and deliver it to users. We validate our method through various experiments and a real-world experiments user study, demonstrating that it achieves a higher success rate. The user study showcases that our approach can generate handover grasps that are more preferred by users. 
    
\end{itemize}
% 1) A vision-based robot-to-human handover system. The robot can track human poses and deliver objects to the receiver in a comfortable way.
% 2) A zero-shot grasp selecting algorithm that take both semantics and geometrics information into consideration to help select optimal grasp in handover. 
% 3) We evaluate our approach in real robot experiments and demonstrate through user surveys that our method enables more user-preferred handovers.

\section{RELATED WORK}

\subsection{Robot-to-Human handover}

Robot-to-human is one of the fundamental tasks in human-robot interaction and has been the subject of research for many years. Through investigation we find that many methods adhere to a similar paradigm: they try to model human grasp region in handover and consider grasping out of it. 

Early on, Aleotti \cite{aleotti_comfortable_2012} observes that humans tend to prefer grasping protruding parts of objects during handovers, such as handles. Thus, he proposes a shape-based approach. By modeling the object and computing its Reeb graph, the method can segment the protruding parts, guiding the robot’s grasp during the handover. This approach leverages the general geometric characteristics of handles but is only applicable to a subset of protruding objects and does not account the object's function. Recent work attempts to acquire such regions through semantic methods, with some directly predicting human hand grasps on objects, while others implicitly represent them using contact maps or affordance detection-based segmentation. Meng \cite{meng_fast_2022} utilizes human grasp prediction to guide the handover grasp selection. This method chooses grasps that are oriented oppositely to the human hand's grasp direction and pre-defines the robot's grasp types. However, the ideal grasp direction may not necessarily be the opposite. Christensen \cite{christensen2022learning} incorporate object semantic information by performing semantic segmentation on the object, while Contact Handover \cite{wang_contacthandover_2024} predicts human contact points on object to guide grasp selection. These works focus on constraining the robot's grasp by region without considering grasp direction.

% These data-driven methods only work within a limited range of objects and semantics, exhibiting poor generalization. 

% and they fail to form a comprehensive understanding of where human and robot should grasp during handover.
\begin{figure*}[htbp]
\centering
\setlength{\abovecaptionskip}{0.cm}
\includegraphics[width = 0.85\linewidth,scale=0.5]{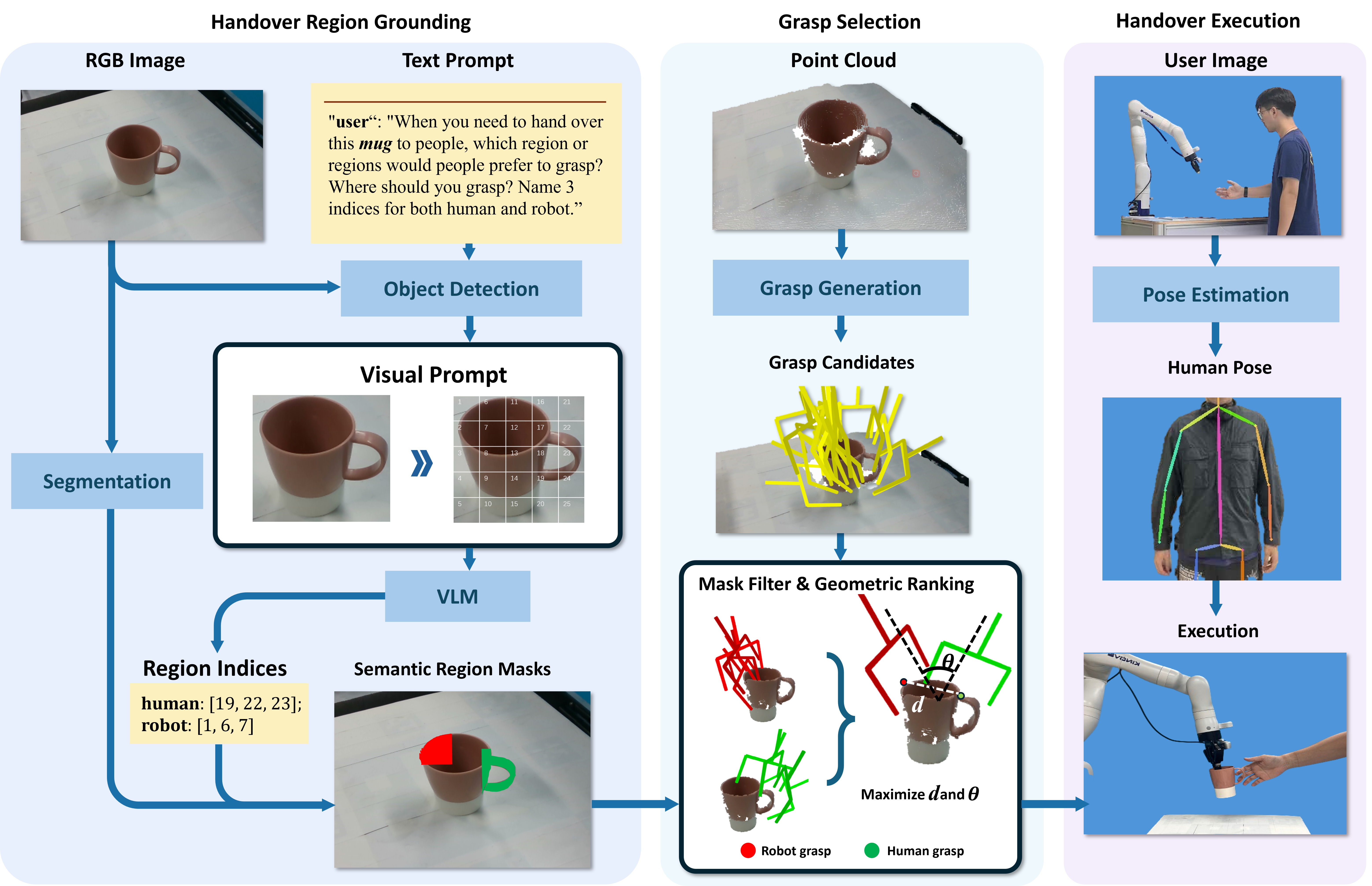}
\caption{Overview of our proposed system. }
\label{figure}
\vspace{-0.1in}
\end{figure*}

\subsection{Foundation Models for Robotic Manipulation}

% Recent advancements in foundation models including Large Language Models (LLMs) and Vision-Language Models (VLMs) have been groundbreaking. Leveraging diverse and extensive training data, these models exhibit remarkable capabilities in open-ended language reasoning and image understanding. In robotics, such capabilities are advantageous for physically grounded tasks, such as tabletop manipulation, navigation, and mobile manipulation. Regarding robot handover, LAN-grasp demonstrates the ability to perform handover-oriented grasps by introducing foudation models. Leveraging an LLM to infer potential human grasping part and VLM to ground it in visual data to enable handover-oriented grasps. While this approach successfully incorporates the strong generalization capabilities of foundation models, it depends on explicit language descriptions of object parts and struggles when part-level information is insufficient for precise handover grasp generation. Although more advanced VLMs, such as GPT-4V, possess stronger vision-language understanding capabilities, their fine-grained visual grounding abilities remain limited. For example, it cannot directly predict the pixel coordinates of grasping points during a handover based on an object's image.

Recent advancements in foundation models, including Large Language Models (LLMs) and Vision-Language Models (VLMs), have been truly transformative. Utilizing extensive and varied training data, these models demonstrate exceptional capabilities in open-ended language reasoning and image comprehension \cite{brown2020language, tang2024foundationgrasp}. In robotics, these strengths are particularly beneficial for tasks involving HRI. Regarding robot handover, LAN-grasp \cite{mirjalili2023lan} demonstrates the ability to perform handover-oriented grasps by using foundation models, with an LLM inferring potential human grasping regions and a VLM grounding this information in visual data to enable effective grasps. Although this approach effectively harnesses the generalization abilities of foundation models, it relies on explicit language descriptions of object parts and encounters challenges when such part-level information is insufficient for accurate grasp selection. While advanced VLMs like GPT-4V offer enhanced vision-language understanding \cite{achiam2023gpt}, their fine-grained visual grounding capabilities are still developing. For instance, GPT-4V fails to directly predict the grasping points in an image at pixel level in handover scenarios. Inspired by  Liu et al.\cite{liu_moka_2024} by converting a trajectory generation to having the VLM select predefined waypoints in image using Set-of-Mark(SoM) \cite{yang2023setofmark} as visual prompts, we adopt a similar strategy in our method.

\section{METHOD}

% As shown in the Fig. \ref{figure}, our system consists of 3 modules. In handover region grounding module, our system figures out where human prefer to grasp when handover happens and corresponding robot grasp region.Then in grasp selection module, the system generates grasp candidates and selects suitable grasp for handover. Then in the delivery module, the system calculates an optimal handover pose which aims to minimize human efforts and execute the grasp and subsequent delivery according to human pose. Detailed discussion follows. 

% To achieve a more effective and generalizable handover, our method focuses on constraining the robot's grasp position and direction based on the handover itself. We propose a method that selects grasp regions using semantic information of the object from VLMs, combined with geometric information to select the grasp. Once the grasp is determined, we compute the optimal handover points based on the human pose and execute it.

To achieve a more effective and zero-shot handover, our method constrains the robot's grasp position and direction according to the handover requirements. We propose a method that selects grasp regions by combining semantic information from VLMs with geometric data and incorporates visual prompts to refine these regions beyond the direct VLM results. After determining the grasp, we compute the optimal handover points based on the human pose and execute the handover.

\subsection{Handover Region Grounding Module} 
A successful handover depends on a proper grasp that concerns human grasp and object function afterward. This module is responsible for identifying suitable grasp regions on the object for robot-to-human handover based on the input object image. To form a comprehensive understanding of the interaction on object between the human hand and the robot, we predict regions for both. Here, we leverage the semantic knowledge of object class and function from a VLM to decide where the human and robot grasp the object during a successful handover. The regions are segmented as image masks for further grasp selection. The detailed implementation process is as follows:

Object RGB image, along with the user's request regarding the handover object, are fed into the object detection module as Fig. \ref{figure}. To achieve a wider range of objects, we employ an open-vocabulary object detector named OWL-ViT \cite{minderer2022simple} to detect and localize any object. Then we prompt it to a VLM for the desired regions. Since current VLMs are not yet capable of directly reasoning about scenes at the pixel level. Inspired by the concept of Set-of-Mark \cite{yang2023setofmark}, speakable labels in the image can significantly enhance the spatial understanding of VLMs. To get a finer understanding of the object's part, we adopt a grid-like label strategy as a visual prompt to enhance the VLM, which specifically in this work is GPT-4V. The specific approach is as follows: after obtaining the object's bounding box, we divide this bounding box evenly into $N^2$ small grids to obtain a finer-grained segmentation and description of the specific parts of the object. Each grid is assigned a numerical label to facilitate understanding by the VLM. An example is illustrated in Fig. \ref{figure}(b). Here, $N$ is set to 5. Coupled with text prompts, this allows us to guide the VLM to output regions by naming grid indices. We designed our prompts to guide the VLM in predicting the grid index corresponding to potential grasp locations for both the human hand and the robot. The content of the prompt is as follows, where \textit{<object>} is the object queried by the user: 

\noindent \textit{"role": "user", "content": "You are an intelligent service robot."}

\noindent \textit{"role": "user", "content": "When you need to hand over this <object> to people, which region would people grasp? And which region would you grasp? Name 3 indices in a format like: $ human:[id_1, id_2, id_3], robot:[id_1, id_2, id_3]$ indicating human and robot grasp regions."}

% Then VLM takes in the above prompt along with the object image overlaid with grids as Fig. \ref{region}(b). It outputs grid indices for two regions. These grids are then merged to form the final bounding boxes for the two grasp regions. However, since the bounding boxes may include non-object parts of the scene, further segmentation is required to isolate the object within these regions for better filtering in the grasp selection process.
% For this, we employ SAM, a promptable segmentation model, to obtain the object’s mask named in text prompt. Finally, we crop the object mask using the parts' bounding boxes obtained in the previous step, yielding semantic masks that contain only the object’s grasp regions: $M_{robot}$ and $M_{human}$ representing the robot and human grasp masks, respectively.

The VLM processes the provided prompt along with the object image overlaid with grids, as shown in Fig. \ref{figure}(b). It then outputs grid indices corresponding to two regions in format \textit{human:$[id_1, id_2, id_3]$, robot:$[id_1, id_2, id_3]$}. These grids corresponding to indices are subsequently merged to have a comprehensive view of the grasp regions for both human and robot. Considering the bounding boxes may encompass non-object areas within the scene, additional segmentation is necessary to isolate the object within these regions for more accurate filtering in the grasp selection process.

To fix this, we adopt Segment Anything Model (SAM) \cite{kirillov2023segany}, a promptable segmentation model, to generate the object mask specified in the text prompt. We then crop the object mask using the bounding boxes obtained in the previous step, resulting in semantic masks that exclusively contain the object’s regions denoted as $M_{robot}$ and $M_{human}$, which represent the robot and human grasp masks, respectively.

\subsection{Grasp Selection Module}
After obtaining the region masks where the human and robot grips occur, we select the final robot grasp suitable for the handover task based on these region masks and geometric-based ranking criteria associated with the handover process. First, we generate diverse grasps on the object. The object's RGB-D image is converted into point cloud data and input into a grasp generation network. Our goal is to generate a diverse set of grasp candidates to enhance the likelihood of achieving a suitable grasp for handover. To achieve this, we utilize the pre-trained model Contact-GraspNet \cite{sundermeyer2021contact}. Contact-GraspNet takes partial point cloud observations of a scene and predicts a 6-DoF robotic gripper pose. 

% To achieve this, we utilize a pre-trained model, Contact-GraspNet. Contact-GraspNet is a U-shaped model based on PointNet++, which takes partial point cloud observations of a scene and predicts 6-DoF robotic gripper pose. Considering in R2H object handover, it is preferable for the robot to minimally intrude into the human's workspace. Following consideration about human-aware grasp in cograsp, we regard that a pair of robotic and human grasps with a large Euclidean distance and a large approach angle as ideal. Since we do not explicitly predict a human grasp, we use the average coordinate of the point cloud within m h to represent the location where the human grasp occurs.

In the context of R2H object handover, it is preferable for the robot to avoid being unnecessarily close to the human. For example, a human would not desire the robot to deliver the object in the pose shown in Fig. \ref{head}(c). Considering the human-aware grasp discussed in \cite{keshari_cograsp_2023} scenarios, we define the following geometric-based ranking criteria for handover grasping: robot and human grasps are prioritized based on larger Euclidean distances and wider approach angles, as these factors enhance human-aware grasp. This criteria are used to select the most suitable grasp.

Since we do not explicitly predict human grasps, we refer to \cite{ardon2021affordance} and implicitly offer the receiver the most suitable grasping region. We analyze two sets of grasp configurations: (i) grasps within $M_{human}$ denoted as $G_\text{human}$ and (ii) grasps within $M_\text{robot}$ denoted as $G_\text{robot}$. The objective is to select a robot grasp in $G_\text{robot}$ that maximizes the distance and approach angle to the grasps in $G_\text{human}$. 

% We conduct clustering on $G_\text{human}$. The centroid of the largest cluster is used as a reference to represent the human grasp. Similar as \{affordance-aware\} , we want to leave the object’s part unoccluded by robot grasp.

All grasp candidates on object are denoted as $G=\left\{g_\text{i} \mid i=1,2, \ldots, n\right\}$ containing $n$ grasp configurations. The point cloud of robot and human grasp regions filtered by  $M_{robot}$ and $M_{human}$ are $PC_{robot}$, $PC_{human}$. As defined in Contact-GraspNet each grasp configuration $g_\text{i}$ is a homogeneous transformation and can be written as a rotation and translation as 
$ g_\text{i}=\begin{bmatrix}
    \mathbf{R_\text{i}} & \mathbf{T_\text{i}}\\
    \mathbf{0} & 1
\end{bmatrix}, \quad \mathbf{R_\text{i}} \in \mathbb{R}^{3 \times 3}, \mathbf{T_\text{i}} \in \mathbb{R}^3 \vphantom{\begin{matrix} 1 \\ 1\\1 \end{matrix}}$

% \[F = \begin{bmatrix}
% A & B \\
% 0 & D
% \end{bmatrix}\]
\noindent which gives the grasp point $\mathbf{T_\text{i}}$, then we filter grasps within these two point clouds by calculating distance as follows:

\begin{equation}
\centering
\setlength{\abovedisplayskip}{1pt}
\begin{aligned}
     &G_\text{human}=\left\{g_\text{i} \mid \exists p_\text{j} \in PC_{\text{hand }}, \left\|\mathbf{T_\text{i}}-p_\text{j}\right\|<\epsilon\right\},\\
    &G_\text{robot}=\left\{g_\text{i} \mid \exists p_\text{j} \in PC_{\text {robot }}, \left\|\mathbf{T_\text{i}}-p_\text{j}\right\|<\epsilon\right\}
\end{aligned}
\end{equation}
% \[
% G_\text{robot}=\left\{g_\text{i} \mid \exists p_\text{j} \in PC_{\text {robot }}, \left\|\mathbf{T_\text{i}}-p_\text{j}\right\|<\epsilon\right\}
% \]
% \[G_\text{human}=\left\{g_\text{i} \mid \exists p_\text{j} \in PC_{\text {hand }}, \left\|\mathbf{T_\text{i}}-p_\text{j}\right\|<\epsilon\right\}\]

% \begin{figure}[htbp]
% \centering
% \includegraphics[width = \linewidth,scale=0.5]{region.png}
% \caption{Example of Handover Region Grounding Module.}
% \label{region}
% \end{figure}
 
\noindent where $\epsilon$ is the Euclidean distance threshold in meters. It is set to 0.02 in practice.

We try to find a grasp in $G_\text{robot}$ that maximizes the distance and approach angle to all elements in $G_\text{human}$ as :

\begin{equation}
\setlength{\abovedisplayskip}{1pt}
\begin{array}{r}
% \\[1mm]
g^*=\operatorname{argmax}_{g_\text{i} \in G_{\text {robot }}}\left(\alpha \cdot \frac{1}{|G_{\text{human}|}} \sum_{g_{\text{j}} \in G_\text{human}} d\left(g_\text{i}, g_\text{j}\right)+\right. \vspace{1ex}\\
\\[1mm]
\left.(1-\alpha) \cdot  \frac{1}{|G_{\text{human}|}} \sum_{g_{\text{j}} \in G_\text{robot}} \theta\left(g_\text{i}, g_\text{j}\right)\right)
\end{array}
\end{equation}

$d(g_\text{i}, g_\text{j})$ and $\theta(g_\text{i}, g_\text{j})$ stand for normalized Euclidean distance and approach angle between $g_\text{i}$ and $g_\text{j}$. $d(g_\text{i}, g_\text{j})$ is straight forward get by:
\begin{equation}
    d(g_\text{i}, g_\text{j})=\frac{\|\mathbf{T_\text{i}} - \mathbf{T_\text{j}}\|-\mu_d}{\sigma_d}
\end{equation} 

$\mu_d$ and $\sigma_d$ are mean and standard deviation of the distance.

To calculate $\theta_\text{i}$, we can calculate the angle between the approach direction of two grasps, which is the z-axis of two reference frames represented by $\mathbf{R_\text{i}}$ and $\mathbf{R_\text{j}}$ which are the rotation matrix for $g_\text{i}$ and $g_\text{j}$:
% \begin{equation}
% \begin{aligned}
%     \theta(g_\text{i}, g_\text{j}))&=\frac{\arccos (\mathbf{z}_\text{i} \cdot \mathbf{z}_\text{j})-\mu_\theta}{\sigma_\theta}, \text{where}
    
%   \mathbf{z_\text{i}}&=\mathbf{R_\text{i}}\left[\begin{array}{l}0 \\ 0 \\ 1\end{array}\right], \mathbf{z_\text{j}}=\mathbf{R_\text{j}}\left[\begin{array}{l}0 \\ 0 \\ 1\end{array}\right]
%   \end{aligned}
% \end{equation}
\begin{equation}
\begin{gathered}
\left.\theta\left(g_{\mathrm{i}}, g_{\mathrm{j}}\right)\right)=\frac{\arccos \left(\mathbf{z}_{\mathbf{i}} \cdot \mathbf{z}_{\mathrm{j}}\right)-\mu_\theta}{\sigma_\theta} \text {, where } \\
\quad \mathbf{z}_{\mathrm{i}}=\mathbf{R}_{\mathrm{i}}\left[\begin{array}{l}
0 \\
0 \\
1
\end{array}\right], \mathbf{z}_{\mathrm{j}}=\mathbf{R}_{\mathrm{j}}\left[\begin{array}{l}
0 \\
0 \\
1
\end{array}\right]
\end{gathered}
\end{equation}

$\mu_\theta$ and $\sigma_\theta$ are mean and standard deviation of the approach angle. In practice, $\alpha$ is set to 0.5. We take $g^*$ as final grasp $g_\text{final}$.

\subsection{Handover Execution Module}

After choosing an optimal grasp for handover, the robot execute grasp and deliver the object to a specific handover pose. This position should fully consider the characteristics of the recipient, such as the distance from robot to human and the length of the recipient’s arm, to ensure that the handover position is reachable. First, we perform pose-tracking on the human body. We use a pre-trained AlphaPose model \cite{alphapose} as the pose estimator to record the spatial positions of the shoulder, elbow, and wrist joints to calculate the lengths of the upper arm $l_\text{u}$ and forearm $l_\text{f}$.

We aim to find a position with minimum effort for humans to reach, so we follow the model described in \cite{liu_object_2021} to quantify the ergonomic cost of arm movement. The total cost, combining joint torque and median joint displacement, is calculated as:
% \vspace{-0.1in}
\begin{equation}
\begin{aligned} C_{\text {total }} & =C_{\text {torque }}+C_{\text {disp }} \\ & =\frac{\sum_{\text{j}=1}^2\left(\tau_\text{j}\right)^2}{C_\text{torque, max }}+\frac{\sum_{\text{j=1}}^2\left(\theta_{\operatorname{mid},\text{j}}-\theta_\text{j}\right)^2}{C_\text{displacement, max }}\end{aligned}
\end{equation}

To simplify the calculation, we follow \cite{parastegari_modeling_2017} to use a simplified right-arm model to represent the arm as a 2-degree-of-freedom planar manipulator in the task plane. Two rotating joints are located at the shoulder and elbow indicated by $\theta_1$ and $\theta_2$, with the rotating axis pointing to the right. $(\theta_1, \theta_2)\in [-45^\circ, 180^\circ] \times [-15^\circ, 140^\circ] $ according to \cite{liu_object_2021} . The Task plane is defined as a plane that includes the shoulder joint and the robot’s base point and is perpendicular to the ground. The mass of upper arm and forearm is set to an average value of 2.6kg and 1.8kg as measured in \cite{clauser1969weight}.

Within the Task Plane, we search $\theta_\text{i}$ in a granular of $1^\circ$ for a position that minimizes the cost $C_\text{total}$ and then calculate $(x,y,z)$ coordinate by forward kinematics of the arm. 
% Regarding orientation, we directly choose a direction $\textbf{v}$ pointing from the center $PC_{robot}$ to the center of $PC_{human}$.

\section{EXPERIMENTS}

% \subsection{Experiment Setup}

We evaluate our method on various daily objects and tools. We first perform an ablation study on handover grasp selection to evaluate the contribution of different components of our method. Subsequently, we conduct handover experiments on a real robot to compare our approach with other baseline methods. Additionally, we design a user study to validate that users prefer the grasps selected by our method over other methods.

\subsection{Ablation Study}

 Our method incorporates both region masks and geometric-based ranking criteria in selecting a grasp for handover; therefore, we conduct separate ablation studies for each. We conduct a dataset consisting of 15 categories of household items and tools with 2 instances in each. Detailed names are provided in the supplementary material.
 
 \noindent \textbf{Metric.}
 Referring to success defined in Sec. \ref{intro}, A robot grasp is considered successful if it avoids the human grasp region and its approach direction does not intrude into the human workspace. The grasp is evaluated by checking if the majority of the object within the human grasp region lies in the direction from the grasp contact point and approach direction. A threshold is applied to quantify this condition. In practice, for every grasp in $G_\text{robot}$, we randomly sample $N$ points from $PC_{human}$ and determine proportion $\textit{p}$ that lie on the same side of a plane defined by the grasp point and direction, where the direction acts as the normal vector. A high proportion keeps the object in the human grasp region on the opposite side of the robot's motion. We choose $N = 100$ and $\textit{p}>0.8$ is considered a success.

\subsubsection{Ablation study on region grounding}In this ablation, we discuss the impact of region masks constrain on the final grasping outcomes. We argue that the semantic information within these masks contributes to grasp selection during handovers. Baselines are set up as below.

 \textbf{Ablation 1 (A1): No semantic information (without human and robot grasp region).} Masks are only used to determine success but not constrain grasp selection. By applying geometric-based ranking criteria to all grasp candidates, we find a pair of grasps that maximize distance and approach angles. The final grasp is randomly chosen between these two grasps.

 \textbf{Ablation 2 (A2): Partial semantic information (without robot grasp region.} Only the human grasp region is used to constrain grasp selection. The final grasp is selected following geometric base ranking criteria outside the human grasp region. 

 \textbf{Ablation 3 (A3): Full semantic information (with both regions, full version of our method). } Both robot grasp region and predict human grasp region are used to constrain grasp selection. The final grasp is selected following geometric base ranking criteria within the robot grasp region. 

\begin{table}[]
% \Large
% \usepackage{multirow}
\caption{Ablation study on region grounding}
\renewcommand\arraystretch{1.7}
\resizebox{\linewidth}{!}{
\begin{tabular}{lccc}
\toprule
\multirow{2}{*}{Method} & \multicolumn{2}{c}{Ablation Factor}                                      & \multirow{2}{*}{Success Rate} \\ \cline{2-3}
                  & \multicolumn{1}{l}{Human Region} & \multicolumn{1}{l}{Robot Region} &                               \\ \hline
A1:w/o human and robot region                & \XSolidBrush                              & \XSolidBrush                               & 32.7\%                           \\ \hline
A2:w/o robot region                & \Checkmark                              & \XSolidBrush & 81.0\%                           \\ \hline
A3:w/ human and robot region               & \Checkmark                              & \Checkmark                              & \textbf{84.3\%}                           \\ \bottomrule
\end{tabular}
}
\label{table1}
\end{table}

We conduct 10 trials per instance and the results of the ablation experiment are shown in Table \ref{table1}. Both A2 and A3 significantly outperform A1, achieving approximately 50\% improvement. The semantic understanding capability of VLM can effectively help select proper grasps in handover. A3 shows a modest improvement over A2, but pre-determining the robot's grasping region can substantially reduce the number of potential grasps to be evaluated and lower the computational burden. At the same time, it helps avoid extreme cases where excessively large distances lead to sub-optimal choices that fail our assumption on approach angle.

\subsubsection{Ablation study on geometric-based ranking criteria}
Next, we evaluate the effectiveness of geometry-based ranking in the selection of handover grasps. Baselines are set up as below.

\textbf{Ablation 1 (B1): Without geometric-based ranking criteria.}  The VLM predicts both robot grasp region and human grasp region. The grasp is selected randomly in the robot grasp region. 

\textbf{Ablation 2 (B2): With geometric-based ranking criteria (full version of our method). } The VLM predicts both robot grasp region and human grasp region. The final robot grasp is chosen according to geometric basic ranking criteria. 

\begin{table}[]
% \Large
% \usepackage{multirow}
\centering
\caption{Ablation study on geometric-based ranking criteria}
\renewcommand\arraystretch{1.6}
\resizebox{\linewidth}{!}{
\begin{tabular}{lcc}
\toprule
\multirow{2}{*}{Method} & Ablation Factor                         & \multirow{2}{*}{Success Rate} \\ \cline{2-2}
                  & \multicolumn{1}{l}{geometric-based Ranking} &                               \\ \hline
B1:w/o geometric ranking                & \Checkmark                              & 62.3\%                           \\ \hline
B2:w/ geometric ranking                & \XSolidBrush                             & \textbf{84.3\%}                           \\ \bottomrule
\end{tabular}
}
\label{table2}
\vspace{-0.1in}
\end{table}
% \vspace{-0.2in}

The experimental results in Table \ref{table2} show that B2 outperforms B1 by 22\%. This indicates that the geometric-based ranking criteria we designed is effective in selecting the appropriate grasp for handover.

\vspace{-0.12in}
\begin{table}[!hbp]
\caption{Likert Items for Evaluating Robot Handover}
\renewcommand\arraystretch{1.6}
\begin{tabular}{cp{7.7cm}}
\toprule
\multicolumn{1}{c}{} & \multicolumn{1}{c}{\textbf{Likert item}}                                                                              \\ \hline
\textbf{1}           & The robot's grasp position made it easy for me to take the object.                                \\
\textbf{2}           & The robot's grasp position did not interfere with where I wanted to grasp the object.             \\
\textbf{3}           & The robot’s grasp direction allowed me to comfortably receive the object without any awkwardness. \\
\textbf{4}           & The robot’s grasp direction did not interfere with my hand space during the handover.             \\ \bottomrule
\end{tabular}
\vspace{-0.1in}
\label{table3}
\end{table}

\subsection{Comparative Experiments}
We carry out real-world robot handover experiments against baselines. We use a Kinova Gen3 robot equipped with a RealSense D435i camera mounted on its wrist to form an eye-in-hand system. To study the practical applicability from the user perspective, we conduct a user study with 10 participants on 10 test objects. We randomly select five objects each from the UMD dataset \cite{Myers:ICRA15}, which is used to train AffNet-DR, and from novel objects. 

\noindent \textbf{Baselines. }Here we choose two baselines: AffNet-DR \cite{christensen2022learning} and LAN-grasp \cite{mirjalili2023lan}. 

AffNet-DR is a typical affordance segmentation based approach. It predicts task related segmentation of an object image. To derive a handover, it predicts segmentation associated with objects' designed function and chooses within the segmentation. AffNet-DR we use here is trained on the UMD dataset from the original paper. 

LAN-grasp is a foundation model based robot grasp generator. Given the task label of \textit{handover}, LAN-grasp is capable of generating a handover-oriented grasp by prompting task description to LLM. It grounds a 2D bounding box representing the grasp region. Since both methods only predict coarse regions, we randomly choose grasp in the predicted region. 

\begin{figure}[H]
\vspace{-0.1in}
\setlength{\abovecaptionskip}{0.cm}
\centering
\includegraphics[width=0.95\linewidth]{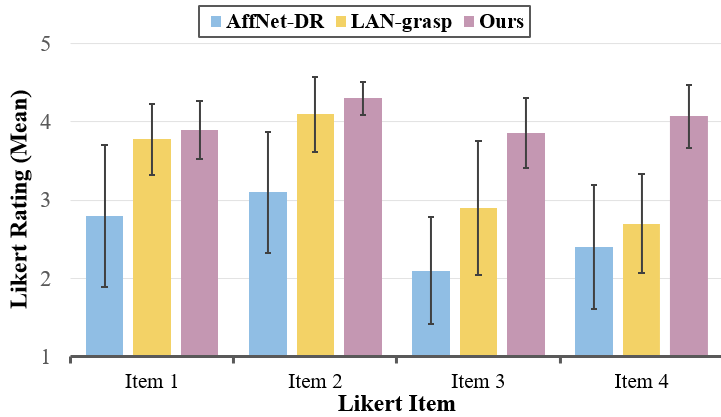}
\caption{Results of our user study in comparison experiments. The horizontal axis represents different items, while the vertical axis shows the average Likert ratings. Three methods are distinguished by different colors. }
\label{graph}
\end{figure}

We count the first successful trial for each object and draw a user study. After each object is tested, participants are invited to fill out a questionnaire with a 5-point Likert scale. The questionnaire is designed to focus on evaluating grasp position and direction as Table \ref{table3}. Each item was measured on a scale from 1 to 5, with 1 representing \textit{Strongly Disagree} and 5 representing \textit{Strongly Agree}. We calculated the mean and standard deviation for each item, and the results are presented in Fig. \ref{graph}.

\begin{figure*}[tp]
% \vspace{-0.1in}
\setlength{\abovecaptionskip}{0.cm}
\centering
\includegraphics[width=1.0\linewidth]{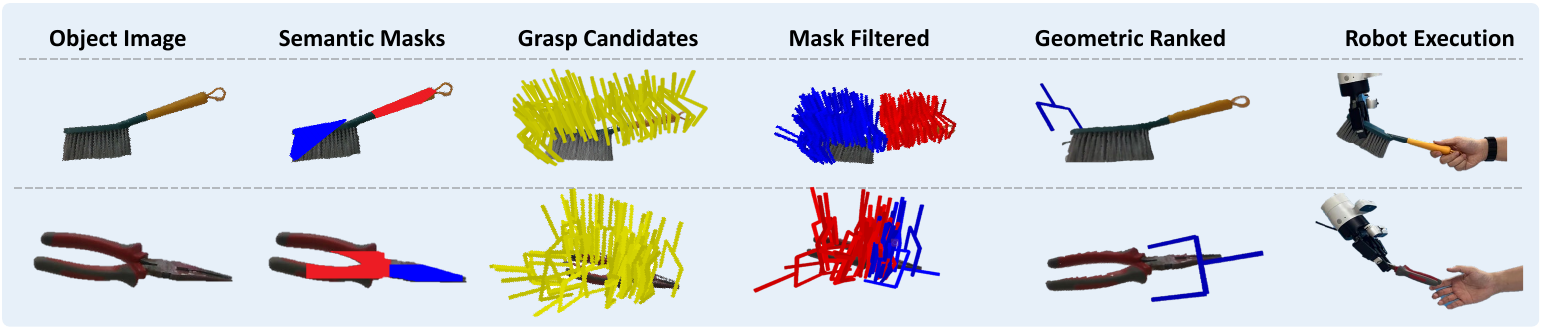}
\caption{Qualitative results of real-world experiments of our method. Each row is a visualization of intermediate results and real-world robot execution is shown in the last column. }
\label{experiment}
\end{figure*}
% \vspace{-0.2in}

Our method outperforms the baselines across all Likert ratings. LAN-grasp performs comparably to our method in the evaluation of the grasping region (items 1 and 2) with consistency to all objects while AffNet-DR receives lower ratings with higher variance, due to its poor generalization on novel objects. Notably, for the two Likert items related to the grasp direction and angle (items 3 and 4), our approach demonstrated superior performance by taking into account the geometric constraints of robotic grasp in handover. This enabled the system to grasp objects in a more user-preferred direction when handing them over to the user.

\begin{figure}[H]
\vspace{-0.1in}
\setlength{\abovecaptionskip}{0.cm}
\centering
\includegraphics[width=1.0\linewidth]{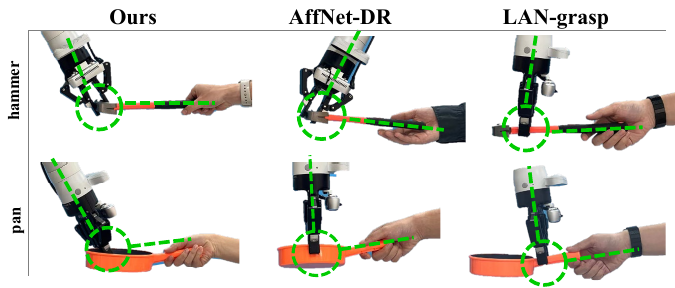}
\caption{Some showcases of real-world robot experiments. Our method can select grasps in proper region with larger approach angles and provide more space in handover compared with other two methods. }
\label{compare}
\end{figure}

\section{CONCLUSIONS}

% In this paper, We propose a zero-shot R2H object handover system. We leverage the semantic knowledge from the VLM to understand objects and ground potential interaction regions on the object. Focusing on the handover itself, we also incorporate geometric constraints on the grasp direction to maximize human ease during the interaction. We conduct ablation studies to demonstrate that both the semantic and geometric information in our method contribute to selecting an appropriate grasp for handover. Real-world experiments demonstrate that our method is capable of performing more human-preferred handover in comparison to other baselines.
In this paper, we propose a zero-shot R2H object handover system. Enhanced by visual prompts, we leverage the semantic knowledge from the VLM to understand objects in finer granularity and ground potential interaction regions on the object. We also incorporate geometric constraints on the grasp direction, focusing on the handover itself to maximize human ease during the interaction. We conduct ablation studies to show that both semantic and geometric information contribute to selecting an appropriate grasp for handover. Real-world experiments demonstrate that our method is capable of performing more human-preferred handover in comparison to other baselines.

\addtolength{\textheight}{-5cm}   % This command serves to balance the column lengths

%%%%%%%%%%%%%%%%%%%%%%%%%%%%%%%%%%%%%%%%%%%%%%%%%%%%%%%%%%%%%%%%%%%%%%%%%%%%%%%%

%%%%%%%%%%%%%%%%%%%%%%%%%%%%%%%%%%%%%%%%%%%%%%%%%%%%%%%%%%%%%%%%%%%%%%%%%%%%%%%%

\bibliographystyle{unsrt}
\bibliography{bibliography}

\end{document}